\documentclass[10pt,twocolumn,letterpaper]{article}

\usepackage{iccv}
\usepackage{times}
\usepackage{epsfig}
\usepackage{graphicx}
\usepackage{amsmath,scalerel}
\usepackage{amssymb}
\usepackage{soul}
\usepackage{xcolor}

\usepackage{lipsum}
\usepackage{comment}
\usepackage{multirow}
\usepackage{makecell}
\usepackage{indentfirst}
\usepackage{subcaption}
\usepackage[accsupp]{axessibility}

\DeclareMathOperator*{\concat}{\scalerel*{\Vert}{\sum}}

\newcommand\da{{$_\downarrow$}}
\newcommand\ua{{$_\uparrow$}}

\usepackage[pagebackref=true,breaklinks=true,letterpaper=true,colorlinks,bookmarks=false]{hyperref}

\iccvfinalcopy 

\def\iccvPaperID{10470} 

\ificcvfinal\fi

\begin{document}

\title{Dense 2D-3D Indoor Prediction with Sound via Aligned Cross-Modal Distillation}

\author{
Heeseung Yun\thanks{Equal Contribution}, \hspace{3pt} Joonil Na,$\!^*$ \hspace{3pt} Gunhee Kim\\
Seoul National University\\
{\tt\small \{heeseung.yun, joonil\}@vision.snu.ac.kr, gunhee@snu.ac.kr}\\
{\tt\small https://github.com/hs-yn/DAPS}
}

\maketitle
\ificcvfinal\thispagestyle{empty}\fi

\begin{abstract}
Sound can convey significant information for spatial reasoning in our daily lives.
To endow deep networks with such ability, we address the challenge of dense indoor prediction with sound in both 2D and 3D via cross-modal knowledge distillation.
In this work, we propose a Spatial Alignment via Matching (SAM) distillation framework that elicits local correspondence between the two modalities in vision-to-audio knowledge transfer.
SAM integrates audio features with visually coherent learnable spatial embeddings to resolve inconsistencies in multiple layers of a student model.
Our approach does not rely on a specific input representation, allowing for flexibility in the input shapes or dimensions without performance degradation.
With a newly curated benchmark named Dense Auditory Prediction of Surroundings (DAPS),
we are the first to tackle dense indoor prediction of omnidirectional surroundings in both 2D and 3D with audio observations.
Specifically, for audio-based depth estimation, semantic segmentation, and challenging 3D scene reconstruction, the proposed distillation framework consistently achieves state-of-the-art performance across various metrics and backbone architectures.
\end{abstract}
\vspace{-10pt}
\section{Introduction}
\label{sec:introduction}

\begin{figure}[t]
    \begin{center}
        \includegraphics[trim=0.0cm 0.0cm 0.0cm 0.0cm,width=0.95\columnwidth]{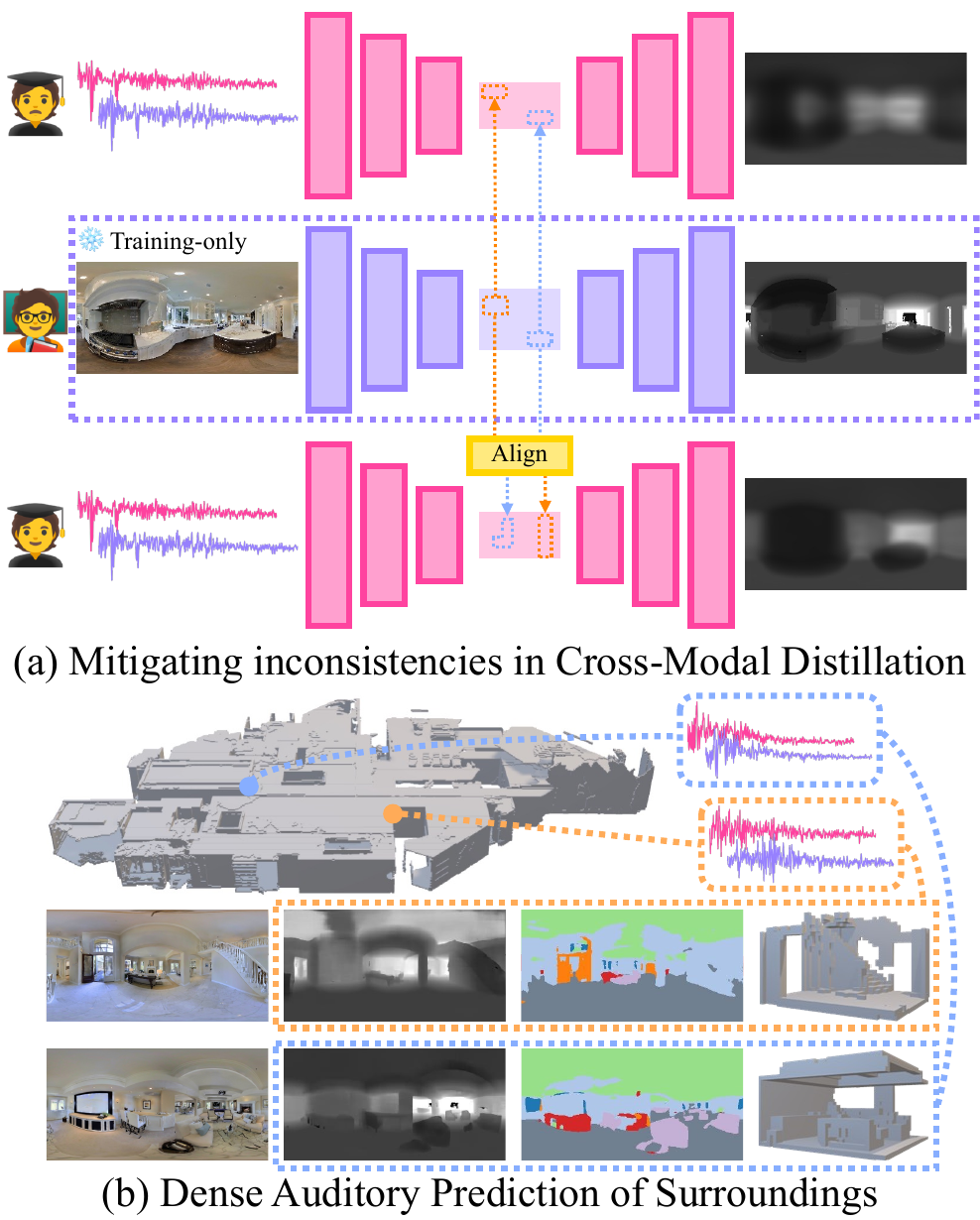}
    \end{center}
    \vspace{-10pt}
    \caption{key idea of our approach. (a) For vision-to-audio cross-modal distillation, instead of direct distillation between geometrically inconsistent modalities, we spatially align the latent feature maps of students with those of teachers. 
      (b) Using auditory input only, we perform three dense predictions of surroundings: depth estimation, semantic segmentation, and 3D scene reconstruction. 
}
    \vspace{-20pt}
    \label{fig:keyidea}
\end{figure}
Humans can get a good grasp of various information about surroundings with hearing without seeing, like the size of a room or the location of an active alarm.
A long line of research has analyzed such intriguing abilities of humans based on interaural differences~\cite{rayleigh1896theory,cherry1953some} or brain activation with respect to spatially aligned audio-visual inputs~\cite{smith2010spatial,cona2019brain}, to list a few.
Accordingly, there is an emerging interest in teaching neural network models for spatial reasoning without seeing.
Such models that spatially perceive the surroundings from sound can be utilized in various environments that are critical for privacy preservation or visually ill-posed (\eg, low illumination or occlusion)~\cite{gan2019self,vasudevan2020semantic,chen2020soundspaces,valverde2021there}.

Since predicting visual properties directly from audio is challenging, cross-modal knowledge distillation~\cite{gupta2016cross} is often utilized, \ie, teaching audio models with the guidance of visual models.
Visual models can make precise predictions about the image of the surroundings, like the location of objects or the depth of a scene.
Thus, using visual models as the teacher, audio models can learn how to predict visual properties in a scene from sound inputs. 
This cross-modal knowledge distillation has been successfully applied to make audio models predict \textit{sparse} attributes, \eg, vehicle tracking~\cite{gan2019self} or indoor navigation~\cite{chen2020soundspaces}.
However, it remains challenging to make \textit{dense} visual predictions about the surroundings from audio.

One of the core challenges behind the dense prediction with audio is to identify fine-grained attributions of the output.
In other words, humans can intuitively make sense of the room layout by hearing, but have difficulty in explaining which bandwidths or timeframes are responsible for their perception.
Unlike distilling an RGB image teacher for a thermal image student that is geometrically consistent up to the pixel level, there is no obvious one-to-one alignment between image and audio.
Hence, it is not feasible to determine which part of the audio spectrogram corresponds to which region of the surrounding.
While using multiple intermediate features of a teacher model as a guide can still be beneficial~\cite{gan2019self,valverde2021there}, it may not be possible to solve the underlying local correspondence problem between the two heterogeneous modalities.

In this work, we are the first to address the dense indoor prediction of omnidirectional surroundings in both 2D and 3D with audio observations.
To resolve the inconsistency problem, we propose a novel Spatial Alignment via Matching (SAM) distillation framework.
SAM matches local correspondences between the two heterogeneous features by making use of learnable spatial embeddings in several layers of the audio student model, combined with loose triplet-based learning objectives.
We retain a set of learnable spatial embeddings to capture spatially varying information of each layer, which are pooled and integrated with initial audio features for alignment.
This allows us to resolve inconsistencies even when the shape of the audio input does not match that of the desired output, making it trivially extendable to a challenging scenario like audio-to-3D distillation.

To comprehensively evaluate the performance of our method, we curate a new benchmark for audio-based dense prediction of surroundings based on Matterport3D~\cite{chang2017matterport3d} and SoundSpaces~\cite{chen2020soundspaces}.
We collect 15.8K indoor scene multimodal observations with task-specific annotations for audio-based depth estimation, semantic segmentation, and 3D scene reconstruction.
In dense auditory prediction tasks spanning from 2D to 3D, our framework consistently improves the performance by a wide margin, which is validated on multiple architectures like U-Net~\cite{ronneberger2015u}, DPT~\cite{ranftl2021vision}, and ConvONet~\cite{Peng2020ConvolutionalON}.
Also, qualitative results demonstrate that our approach can precisely predict the structure of the indoor environment with hearing without seeing.

\section{Related Works}
\label{sec:related_works}

\textbf{Indoor Multimodal Scene Analysis.}
Extensive research has been conducted to understand indoor surroundings for given various inputs.
Using monocular images as input, many visual scene understanding tasks like depth estimation, semantic segmentation, and surface normal estimation have been studied~\cite{silberman2012indoor,chang2017matterport3d,armeni2017joint}.
In addition, 3D-based methods for semantic segmentation, object recognition, and floorplan reconstruction have been proposed with voxel or mesh-based representations~\cite{chang2017matterport3d,armeni2017joint,armeni20163d,dai2017scannet}.
When performing such tasks, combining different modalities as inputs is proven to be effective, such as RGB with depth information for semantic segmentation~\cite{couprie2013indoor} or voxels with point clouds for 3D segmentation~\cite{zhang2020deep}.
Recently, 2D vision-language models are successfully employed for open-vocabulary 3D scene understanding~\cite{ha2022semantic,roh2022languagerefer}.

There has been a surge of interest in combining audio and visual signals to tackle visual or acoustic tasks in indoor environments.
Some prior works generate binaural audio~\cite{gao20192} or scene-aware auditory responses~\cite{chen2022visual,singh2021image2reverb} by utilizing visual surroundings as a reference.
Binaural audio is simulated from a 3D scene for audio-visual embodied navigation~\cite{chen2020soundspaces,chen2022soundspaces}.
Audio signals can help improve performances in visual tasks like floorplan reconstruction~\cite{purushwalkam2021audio} and depth estimation of normal field-of-views~\cite{gao2020visualechoes,parida2021beyond}.

\textbf{Cross-modal Knowledge Distillation.}
Knowledge distillation~\cite{hinton2015distilling} aims at transferring knowledge from a teacher model to a student model by minimizing the distances between the two logit distributions.
Cross-modal distillation~\cite{gupta2016cross} enhances this transfer by ensuring that the intermediate features of the student model align with those of the teacher model when their input modalities are different.
Distillation between different modalities can improve the robustness of prediction under diverse conditions, such as utilizing depth sensors in student models by distilling object detection, action recognition, or semantic segmentation models~\cite{hoffman2016cross,garcia2018modality,jiao2019geometry}.
Likewise, Vobecky \etal~\cite{vobecky2022drive} leverage LiDAR and image inputs to generate spatially consistent object proposals for semantic segmentation.

Cross-modal distillation can be applied to the scenarios where no explicit correspondence exists between the two modalities.
Zhao \etal~\cite{zhao2018through} use a student model with radio signals for human pose estimation via distillation.
Roheda \etal~\cite{roheda2018cross} conditionally utilize noisy observations of available sensors like seismic sensors to enhance image quality.
Also, audio-only and image-only teachers can teach a video-only student model via shared latent embedding~\cite{chen2021distilling} or long short-term memory networks~\cite{gao2020listen} for better classification.
Other examples include knowledge transfer of speech models for visual lip reading~\cite{afouras2020asr,zhao2020hearing} or visual captioning models for audio captioning~\cite{zhao2022connecting}.

\textbf{Spatial Reasoning with Sound.} 
Sound contains valuable information for spatial reasoning.
Embodied agents can navigate indoor environments by relying solely on auditory input~\cite{chen2020soundspaces}, and their exploration behavior can be promoted by referring to auditory feedback~\cite{zhao2022impact}.
Other prior works focus on the spatial localization of audio sources~\cite{yasuda2022echo}, 3D face synthesis from speech~\cite{wu2022cross}, and depth estimation on a robot~\cite{christensen2020batvision,tracy2021catchatter,irie2022co}.
Sound-only models can benefit from the cross-modal distillation of visual teacher models for fine-grained spatial understanding.
Vision-to-audio knowledge distillation has shown compelling performance in vehicle localization~\cite{gan2019self,valverde2021there}, obstacle detection~\cite{chen2022structure}, and collision probability estimation~\cite{raistrick2021collision}.
However, prior works are limited to the sparse prediction of the surrounding environment (\eg, bounding boxes), while the dense prediction remains challenging.

Closest to our approach is Binaural SoundNet~\cite{vasudevan2020semantic,dai2022binaural}, as it improves outdoor dense prediction performance through the cross-modal distillation of multiple tasks.
However, our work has three significant differences.
First, we perform indoor semantic segmentation and 3D scene reconstruction from audio as new dense prediction tasks.
Second, SoundNet does not consider feature-level alignment, while our method hierarchically leverages spatial alignment via matching for fine-grained vision-to-audio distillation.
Finally, instead of designing a new architecture for modeling audio inputs~\cite{gan2019self,dai2022binaural} or forcing specific input representations~\cite{vasudevan2020semantic,valverde2021there}, we take the audio input as is and adapt off-the-shelf vision models for audio-based dense prediction.

\section{Approach}
\label{sec:approach}

\begin{figure}[t]
    \begin{center}
        \includegraphics[trim=0.0cm 0.0cm 0.0cm 0.0cm,width=\columnwidth]{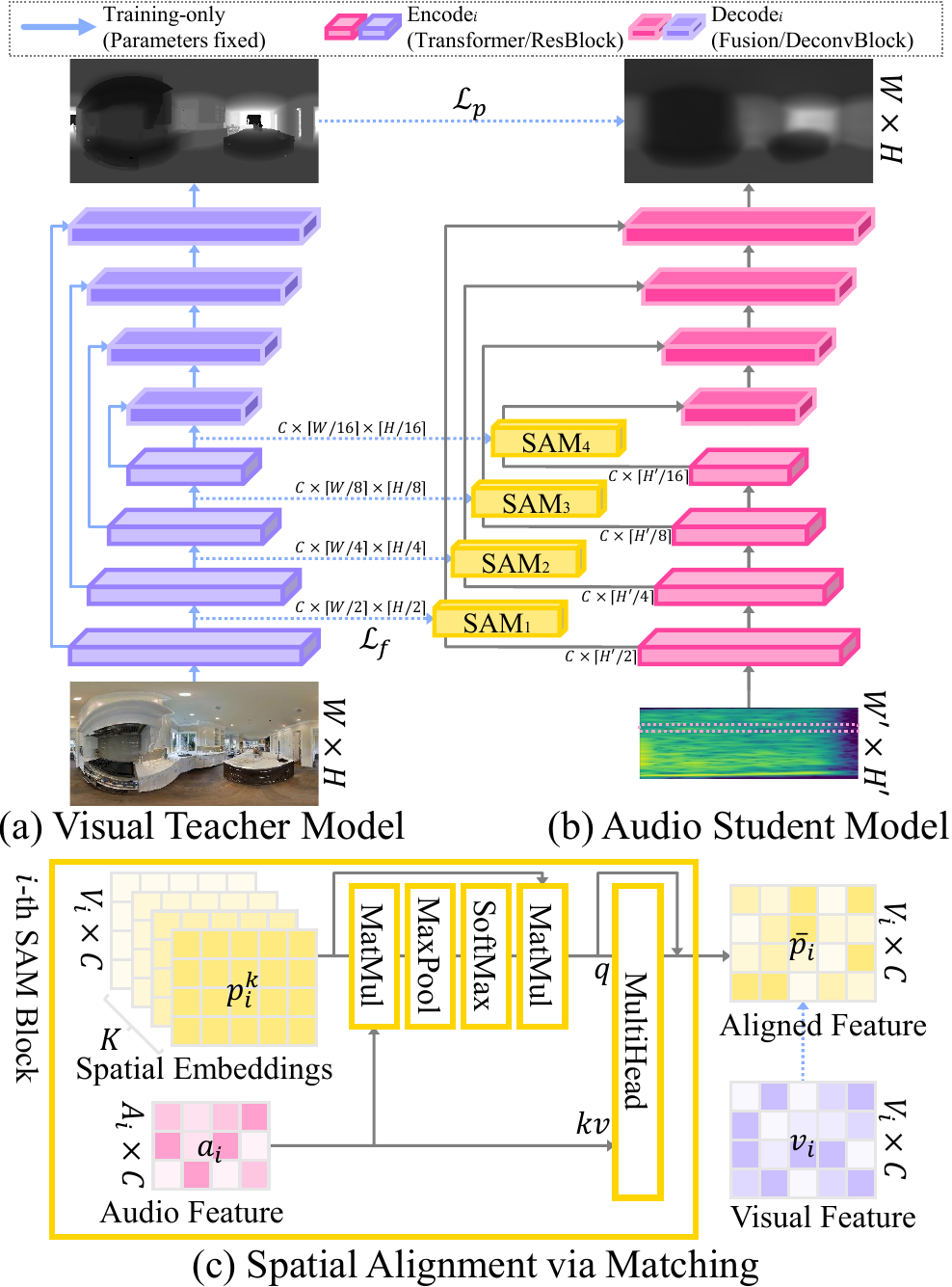}
    \end{center}
    \vspace{-10pt}
    \caption{
Overview of our spatial alignment via matching distillation framework.
}
    \vspace{-10pt}
    \label{fig:architecture}
\end{figure}

Our goal is to predict various dense properties of indoor surroundings without visual input by leveraging binaural audios, \eg, depth, semantic labels, and 3D structures.
To this end, we present a framework for vision-to-audio knowledge distillation that does not rely on specific architecture and entails the alignment of heterogeneous features, as shown in Fig.~\ref{fig:architecture}.
Given a pre-trained visual teacher, we aim to train an audio student model using paired audio-visual observations as training data.

We start by reviewing the basics of vision-to-audio knowledge distillation and the challenges in adapting such methods for dense auditory prediction of surroundings (\S \ref{subsec:crossmodalkd}).
Next, we explain the proposed spatial alignment via matching distillation (\S \ref{subsec:sam}).
Finally, we outline training and inference procedures shared among different tasks (\S \ref{subsec:training}).
Commonly used variables are defined as follows.

\noindent
\begin{minipage}{\columnwidth}
\vspace{6pt}
\begin{minipage}[b]{\columnwidth}
\centering\begin{tabular}{@{\hspace{0.5\tabcolsep}}c@{\hspace{0.5\tabcolsep}}|@{\hspace{1.\tabcolsep}}l@{\hspace{0.5\tabcolsep}}}
\hline
$a_{\text{in}}, v_{\text{in}}$ & Audio, visual input ($\mathbb{R}^{W' \times H' \times 2}, \mathbb{R}^{W \times H \times 3}$) \\
$a_{\text{out}}, v_{\text{out}}$ & Audio, visual prediction output ($\mathbb{R}^{W \times H}$) \\
$a_i, v_i$ & Features at layer $i$ ($\mathbb{R}^{A_i \times C}$, $\mathbb{R}^{V_i \times C}$) \\
$a_i(j),v_i(j)$ & $j$-th feature at layer $i$ ($\mathbb{R}^C$) \\
$A_i,V_i$ & Feature resolution ($w_i^a \times h_i^a,w_i^v \times h_i^v$) \\
$p_i^k$ & $k$-th learnable spatial embedding at layer $i$ \\
         & ($\mathbb{R}^{V_i \times C}$, $0 \leq k < K$) \\
$\bar{p}_i$ & Aligned feature at layer $i$ ($\mathbb{R}^{V_i \times C}$) \\
\hline
\end{tabular}
\end{minipage}
\end{minipage} 

\subsection{Vision-to-Audio Knowledge Distillation}
\label{subsec:crossmodalkd}

Cross-modal distillation from a visual teacher model to an audio model has two significant advantages: 
(i) training without labeled data by turning to the teacher model's prediction (pseudo-GT) and (ii) teaching fine-grained knowledge to the student model via feature distillation.
In general, cross-modal distillation for spatial reasoning leverages both pseudo-GT and feature outputs from one or more layers for fine-grained knowledge transfer~\cite{gupta2016cross}:
\begin{equation}
    \label{eq:ckd_objective}
    \mathcal{L}_{\text{crossKD}} = d(v_\text{out}, a_\text{out}) + \lambda \sum_i\sum_j d(v_i(j), a_i(j)),
\end{equation}
where $d(\cdot,\cdot)$ is a distance function.
This objective is well-defined for two modalities that are consistent up to pixel level (\eg, distilling an RGB teacher to a depth student).
On the other hand, it is less plausible to use the same method for vision-to-audio knowledge distillation.

The main difficulty that hinders knowledge transfer is the semantic and shape inconsistencies of the two heterogeneous modalities.
First, the semantics of audio and visual features are not coherent with each other.
For example, in the second term of Eq.~(\ref{eq:ckd_objective}), the $j$-th feature of an audio-only model at layer $i$ may not always match the corresponding feature of a vision-only model.
This lack of correspondence between the features of the two modalities makes direct distillation depicted in Fig.~\ref{fig:keyidea}-(a) less effective,
which is empirically in line with previous research on vehicle tracking~\cite{gan2019self,valverde2021there}.
Second, the shape of audio input is usually not identical to visual input, and simple interpolation of an audio input often deteriorates the performance.
Moreover, it is even more challenging when the dimensions of the two modalities do not match, \eg, predicting 3D surroundings from audio.
Hence, it is necessary to establish a method that can effectively align with visual features regardless of specific input shapes other than na\"ive resizing or cropping.

\subsection{Spatial Alignment via Matching}
\label{subsec:sam}

To resolve the challenges mentioned above, we introduce a novel method for cross-modal knowledge distillation of two heterogeneous modalities without semantic and shape consistency.
We coin this method Spatial Alignment via Matching (SAM), which comprises three major components: input representation, learnable spatial embeddings, and feature refinement.
To obtain the spatially aligned features for the $i$-th layer of the audio encoder, we can allocate a SAM block that accounts for both feature alignment and shape discrepancy, \ie, SAM$_i: \mathbb{R}^{A_i \times C} \rightarrow \mathbb{R}^{V_i \times C}$.

\textbf{Input Representation.}
Using Short-Term Fourier Transform (STFT) spectrograms of raw binaural audios, we can exploit any 2D deep networks as commonly done in audio representation learning~\cite{hershey2017cnn,gong2021ast}.
However, unlike previous works that rely on pseudo-GT~\cite{vasudevan2020semantic,dai2022binaural} or require identical shapes for feature-level distillation~\cite{gan2019self,valverde2021there}, our method can be trivially applied where $(w_i^a,h_i^a) \neq (w_i^v, h_i^v)$.

In addition, SAM can handle more challenging scenarios like 1D encoders, \ie, $w_i^a=1$ or $h_i^a=1$, by regarding the input spectrogram as a set of 1D patches.
Decomposing the spectrogram into time bands ($W'\times 1$) or frequency bands ($1\times H'$) can effectively reduce the feature shape and replace 2D with 1D operations.
This allows for more efficient encoder implementation in terms of memory and time, making it applicable to memory-intensive scenarios.

\textbf{Learnable Spatial Embeddings.}
It is essential to retain features that are spatially well-aligned with dense prediction output, especially when the input is not aligned with the output modality.
In this regard, we design learnable spatial embeddings as a container to capture spatially varying information in paired audio-visual observations.
We maintain a set of embeddings $p_i^0,...,p_i^{K-1}$ identical in shape with visual features for each SAM and transform the shape of student features before the decoder.
The number of learnable embeddings $K$ may vary across layers, where more slots can be assigned to reconstruct high-level features.

For $K$ learnable embeddings, we first derive a similarity matrix $T_i \hspace{-1.6pt}\in\hspace{-1.6pt} \mathbb{R}^{K \times V_i}$, which represents the proximity between provided audio feature $a_i$ and the $k$-th spatial embedding.
We compute the pairwise similarity between the $j$-th audio feature and the $l$-th feature in a spatial embedding, \ie, $a_i(j), p_i^k(l)\hspace{-1.6pt} \in \hspace{-1.6pt}\mathbb{R}^C$, and select the maximum value along the $j$ dimension:
\begin{equation}
    \label{eq:audiopool}
    T_i = \concat_{k=0}^{K-1} T_i^k = \concat_{k=0}^{K-1} \max_{j} p_i^kW_ia_i(j),  
\end{equation}
where $W_i \in \mathbb{R}^{C \times C}$ is a linear projection and $||$ is a concatenation operator.
That is, higher similarity implies more coherency between the audio features and spatial embeddings at each region, allowing us to obtain features that are spatially aligned with the visual features.


By applying softmax along the $K$ dimension of similarity matrix $T_i$, we then obtain a pooled embedding $\hat{p}_i \in \mathbb{R}^{V_i \times C}$ as a linear combination of embeddings:

\begin{equation}
    \label{eq:embedpool}
    \hat{p}_i = \concat_{l=0}^{V_i-1} \sum_{k=0}^{K-1} \frac{e^{T_i^k(l)}}{\sum_k e^{T_i^k(l)}} p_i^k(l).
\end{equation}
The softmax term can be interpreted as a probability distribution of selecting $k$-th embedding for high audio-visual correspondence, making $\hat{p}_i$ coherent with audio features while maintaining the spatial structure of visual features.

\textbf{Refinement with Student Features.}
For better coherence with audio features, we refine the pooled embedding $\hat{p}_i$ using audio feature $a_i$ as keys and values by leveraging a multi-head attention mechanism (MultiHead)~\cite{vaswani2017attention}:
\begin{equation}
    \label{eq:multihead}
    \bar{p}_i = \text{MultiHead}(\hat{p}_i, a_i, a_i)+\hat{p}_i.
\end{equation}

As a result, we obtain the aligned feature $\bar{p}_i$ from the SAM block at layer $i$.
SAM can facilitate the spatial alignment between features at one (\ie, a bottleneck between the encoder and decoder) or more layers.
For instance, it can be applied to the global residual connection in pyramid-like architectures~\cite{ronneberger2015u,lin2017feature,zhao2017pyramid} to ensure shape consistency, as depicted in Fig.~\ref{fig:architecture}--(a-b).

\subsection{Training and Inference}
\label{subsec:training}

\textbf{Network Architecture.}
For teacher models in each task, we follow the training procedure established in previous literature~\cite{ranftl2021vision,zhao2017pyramid,Peng2020ConvolutionalON}.
For simplicity, we train the teacher models using ground truth labels in the training split, while we also report the cross-modal distillation performance of non-iid settings in Appendix.
We use ImageNet~\cite{deng2009imagenet} pre-trained weights for training teacher models in 2D tasks.
Trained teacher models are only utilized during the training of a student model, with parameters fixed.
 

Our approach can be applied to a wide range of architectures for dense auditory prediction.
We demonstrate this by using U-Net~\cite{ronneberger2015u} with a ResNet-50~\cite{he2016deep} backbone and Dense Prediction Transformers (DPT)~\cite{ranftl2021vision} with a ViT-B/16~\cite{dosovitskiy2021image} backbone as representative examples of convolutional networks and vision transformer variants, respectively.
We exploit Convolutional Occupancy Networks (ConvONet)~\cite{Peng2020ConvolutionalON} as a base architecture for 3D reconstruction.
Using paired audio-visual observations, student models are trained to mimic the output of the teacher model.

\textbf{Learning Objective.}
We minimize the task-specific distance metric between the student and teacher model's prediction (pseudo-GT), \ie, $\mathcal{L}_p=d(v_\text{out}, a_\text{out})$.
To facilitate the cross-modal distillation, 
we integrate an auxiliary feature loss that promotes local coherence between $a_i$ and $v_i$ by optimizing the distance among triplets $(v_i(j), a_i(k), a_i(k'))$:
\begin{equation}
    \label{eq:loss_triplet}
    \mathcal{L}_f^i = \frac{1}{V_i} \hspace{-1pt} \sum_j \hspace{-2pt} \sum_{k' \in \mathcal{N}_k} \hspace{-4pt} \max(0, m - v_i(j) *a_i(k) + v_i(j)*a_i(k')),
\end{equation}
where $m=0.3$ is a margin, $\mathcal{N}_k$ is a set of negative samples regarding $a_i(k)$, and $*$ indicates cosine similarity.
Since there are no ground truth positive pairs for local correspondence, we use $a_i(k)=\arg\max_{a_i(k)}a_i(k)*v_i(j)$ as a loosely defined positive pair.
For $\mathcal{N}_k$, we either deem all the other features in $a_i$ as negative or randomly select one among adjacent features, depending on the convergence of feature loss.
In summary, our learning objective is as follows:
\begin{equation}
    \label{eq:loss_ours}
    \mathcal{L}_\text{Ours} = \mathcal{L}_p + \lambda \sum_i \mathcal{L}_f^i,
\end{equation}
where $\lambda$ is a task-specific hyperparameter to balance the scale between the pseudo-GT loss and feature loss.
We use up to four SAM blocks for all experiments, where we set $K=64$ for the last SAM (SAM$_4$) and reduce the number by a factor of four.
We train the student model from scratch, and during inference, we do not use any input, feature maps, or modules related to the visual modality; only the audio input and the trained audio-only student model are utilized.
Further details are deferred to Appendix.

\begin{figure*}
\begin{minipage}[b]{0.64\textwidth}
\centering
\begin{tabular}{cl|ccccc}
\hline
&& MAE\da & RMSE\da & $\delta_1$\ua & $\delta_2$\ua & $\delta_3$\ua \\ \hline
& Teacher~\cite{ronneberger2015u} & 0.6524 & 1.1296 & 0.7633 & 0.8966 & 0.9328 \\ \hline
& BilinearCoAttn~\cite{irie2022co}             & 1.2101 & 1.8366 & 0.5128 & 0.7009 & 0.8139 \\
& BatVision~\cite{christensen2020batvision}    & 0.9345 & 1.5740 & 0.6284 & 0.7975 & 0.8806 \\  
& MM-DistillNet~\cite{valverde2021there}       & 0.8995 & 1.5812 & 0.6633 & 0.8178 & 0.8902 \\  \hline
\multirow{7}{*}{\rotatebox[origin=c]{270}{\makecell{V$\rightarrow$A \\ U-Net \cite{ronneberger2015u}}}}
& Pseudo-GT ($\mathcal{L}_p$)~\cite{vasudevan2020semantic} & 0.9572 & 1.6436 & 0.6258 & 0.7971 & 0.8771 \\
& + Rank~\cite{gan2019self}         & 0.9524 & 1.6350 & 0.6279 & 0.7986 & 0.8786 \\
& + MTA~\cite{valverde2021there}    & 0.9572 & 1.6392 & 0.6243 & 0.7956 & 0.8782 \\
& + SAM$_\text{MultiHead}$          & 0.8789 & 1.5604 & 0.6774 & 0.8256 & 0.8955 \\
& + SAM$_\text{SpatialEmbeddings}$  & 0.8760 & 1.5468 & 0.6787 & 0.8267 & 0.8965 \\
& + SAM$_{3,4(K=1)}$                & 0.8704 & 1.5467 & 0.6857 & 0.8302 & 0.8978 \\ 
& + \textbf{SAM$_{3,4}$}            & \textbf{0.8633} & \textbf{1.5397} & \textbf{0.6869} & \textbf{0.8308} & \textbf{0.8982} \\ \hline
\multirow{6}{*}{\rotatebox[origin=c]{270}{\makecell{V$\rightarrow$A \\ DPT \cite{ranftl2021vision}}}}
& Pseudo-GT ($\mathcal{L}_p$)~\cite{vasudevan2020semantic} & 0.8926 & 1.5851 & 0.6684 & 0.8243 & 0.8943 \\
& + Rank~\cite{gan2019self}         & 0.9130 & 1.6017 & 0.6607 & 0.8159 & 0.8869 \\
& + MTA~\cite{valverde2021there}    & 0.8913 & 1.5819 & 0.6694 & 0.8263 & 0.8953 \\
& \textbf{+ SAM$_{4}$}              & 0.8517 & \textbf{1.5276} & 0.6971 & 0.8344 & 0.8986 \\
& \textbf{+ SAM$_{3,4}$}            & \textbf{0.8443} & 1.5351 & \textbf{0.7019} & \textbf{0.8392} & 0.9000 \\
& \textbf{+ SAM$_{1,2,3,4}$}        & 0.8497 & 1.5346 & 0.6992 & 0.8380 & \textbf{0.9002} \\\hline
\end{tabular}
\captionof{table}{Comparison of depth estimation accuracy on DAPS-Depth test split.}   
\label{tab:depth_full}
\end{minipage}
\hfill
\begin{minipage}[b]{0.3\textwidth}
    \centering
    \includegraphics[trim=0.0cm 0.0cm 0cm 0.0cm,clip,width=0.96\textwidth]{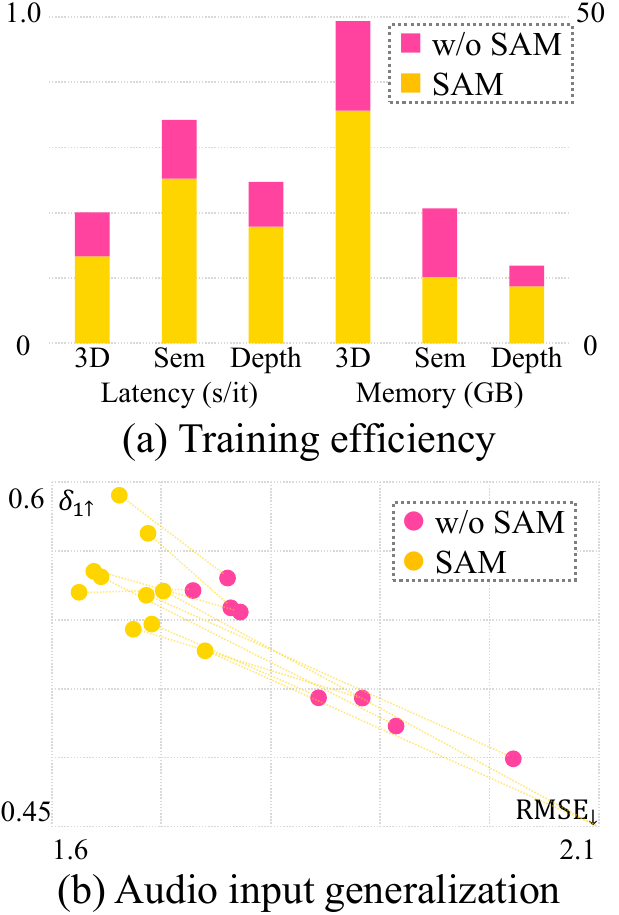}
    \captionof{figure}{Analysis on distillation efficiency and input generalization.}
    \label{fig:efficiency}
\end{minipage}
\vspace{-10pt}
\end{figure*}

\section{Experiments}
\label{sec:experiments}

We first discuss a new benchmark for three audio-based dense prediction tasks of scene understanding (\S \ref{subsec:DAPS}).
We then present the results of our approach for audio-based depth estimation, semantic segmentation, and 3D scene reconstruction tasks (\S \ref{subsec:depth}--\ref{subsec:3d}).

\subsection{The DAPS Benchmark}
\label{subsec:DAPS}

To evaluate the 2D and 3D dense prediction performance with audio, both the audio signal and the information regarding its surrounding space are required.
Since none of the existing works benchmark multifaceted aspects of the omnidirectional surroundings as a whole, we organize a new benchmark upon existing simulators and datasets.
We coin this benchmark Dense Auditory Prediction of Surroundings (DAPS).
DAPS comprises 15.8K indoor scene observations with labels, where each sample consists of binaural audio, RGB panorama, and 3D voxel triples as observation and dense labels for three different tasks, as illustrated in Fig.~\ref{fig:keyidea}-(b).

SoundSpaces~\cite{chen2020soundspaces} can simulate sound in indoor environments; for example, it includes Matterport3D~\cite{chang2017matterport3d} that deals with the material properties and layouts of a scene. 
Once setting the position and orientation of the recording agent in SoundSpaces, we obtain the recordings with respect to a set of emitter and receiver coordinate pairs.
For simplicity, we report the results when the coordinates of an emitter and a receiver are identical.

After sampling coordinates information, we employ the Habitat simulator~\cite{habitat19iccv} to extract multimodal observations of a scene.
We obtain RGB, depth, and semantic labels in equirectangular format from each location.
To further collect 3D information of a scene, we extract the meshes surrounding the specified coordinate by truncating them, \ie, 2.5m$\times$2.5m$\times$2m.
Then, we use clustering-based filtering to remove noisy groups of meshes and keep only the most salient components.
Finally, we generate 3D voxels from meshes for 3D reconstruction.

We carefully exclude the samples with weak auditory signals, such as outdoor scenes with high levels of noise, to maintain the quality of the benchmark.
Specifically, for 2D dense prediction tasks, we eliminate samples whose labels have more than 10\% missing pixels or noisy annotations.
For 3D dense prediction, we exclude the samples with corrupted voxels by selecting the 95\% lower confidence bound of the number of occupied voxels. 
We use 11.6K samples for training, 1.6K samples for validation, and 2.6K samples for testing in all experiments.

\subsection{Results of Depth Estimation}
\label{subsec:depth}

\subsubsection{Experiment Settings}
Following previous works on depth estimation~\cite{ranftl2021vision,jiang2021unifuse}, we predict the depth of the whole surroundings given binaural audio from the scene.
We follow the decoder design of \cite{jiang2021unifuse} to train the model with the Inverse Huber loss.
We report the results of sinusoidal sweep-convolved binaural inputs following the convention of \cite{gao2020visualechoes,parida2021beyond,christensen2020batvision,irie2022co}.
We also report the results of natural audio inputs~\cite{chen2020soundspaces} in Fig.~\ref{fig:efficiency}-(b).

\textbf{Evaluation Metrics.}
We report the mean absolute error (MAE), root mean squared error (RMSE), and delta accuracy ($\delta_1, \delta_2, \delta_3$) for evaluation.
MAE and RMSE reflect the error rate of our prediction, while the delta accuracy indicates the relative correctness of our prediction, \ie, $\max(\frac{a_{\text{out}}}{v_{\text{out}}}, \frac{v_{\text{out}}}{a_{\text{out}}}) < 1.25^i$.
To demonstrate the efficiency of our approach, we also report the memory allocation on GPU and latency during training.

\textbf{Baselines.}
We include some state-of-the-art audio-only and distillation models as baselines~\cite{christensen2020batvision,valverde2021there,irie2022co}, which are originally designed to predict bounding boxes or depth maps from a normal field-of-view with multi-channel audios.
We also report the performance of losses proposed in \cite{vasudevan2020semantic,gan2019self,valverde2021there} combined with U-Net or DPT for fair comparison.

\begin{table}[t]
\centering
\begin{tabular}{l|ccc}
\hline
                            & MAE\da & RMSE\da & $\delta_1$\ua \\ \hline
Mono                        & 1.0783 & 1.7543 & 0.5829 \\ \hline
$16\times16$ Patch          & 0.8903 & 1.5786 & 0.6753 \\
$1\times H'$ Patch (freq.)  & 0.8902 & 1.5607 & 0.6629 \\ 
$W'\times 1$ Patch (time)   & \textbf{0.8497} & \textbf{1.5346} & \textbf{0.6992} \\\hline
Embeddings$_\text{NonSpatial}$ & 0.8777 & 1.5334 & 0.6757 \\
Embeddings$_\text{Oracle}$  & 0.5622 & 1.0308 & 0.8156 \\ \hline
\end{tabular}
\caption{Influence of input representation and learnable spatial embeddings in DPT+SAM on DAPS-Depth test split.}
\label{tab:depth_ablation}
\vspace{-10pt}
\end{table}

\subsubsection{Results and Analyses}

\textbf{Comparison with Prior Arts.}
Table~\ref{tab:depth_full} summarizes the accuracy on DAPS-Depth test split.
Compared to previous works on audio-only and distillation-based auditory depth estimation, our method achieves significant performance boosts across all metrics.
For both U-Net and DPT, directly minimizing the feature distance between the teacher and the student (\ie, +Rank/MTA) contributes marginally to the performance.
Instead, adopting the proposed spatial alignment via matching improves the performance substantially, up to 10\% (MAE) for U-Net.
It is also worth noting that U-Net with SAM displays comparable performance with DPT variants.
One of the important aspects of our approach is its efficiency, as illustrated in Fig.~\ref{fig:efficiency}-(a).
Compared to previous distillation methods, DPT+SAM improves both time and space efficiency by 27\%, where the gap becomes wider for the other two tasks.

\textbf{Ablation Studies.}
In Table~\ref{tab:depth_full}, replacing full SAM blocks with multi-head attention (SAM$_\text{MultiHead}$) or learnable spatial embeddings (SAM$_\text{SpatialEmbeddings}$) deteriorates the absolute error rate by 1.5-1.8\%.
Reducing the number of spatial embeddings per layer to one (SAM$_{3,4 (K=1)}$) is also harmful to performance.
Increasing the number of SAM blocks for alignment can be beneficial, but forcefully matching the low-level vision features with audio features (\ie, SAM$_{1,2}$) does not improve the prediction accuracy.

Table~\ref{tab:depth_ablation} analyzes the influence of different patch designs and spatial embeddings.
Both frequency and time patches are more efficient than the regular patch, but only the time patch introduces significant performance gain.
This implies that aggregating all frequency responses per short time span is a preferred input representation for dense auditory prediction.
Also, the degraded performance of $\mathbb{R}^{K\times 1\times C}$ spatial embeddings instead of $\mathbb{R}^{K\times V_i \times C}$ (\ie, non-spatial embeddings) stresses the importance of securing spatially varying information for matching.
Finally, using actual visual features instead of learnable embeddings (\ie, oracle embeddings) displays on-par performance with the teacher model.

\textbf{Generalization to Natural Audio Inputs.}
Fig.~\ref{fig:efficiency}-(b) reports the distillation performance of U-Net trained with diverse audio samples randomly selected from \cite{chen2020soundspaces}.
Not only our approach consistently achieves better performance, but the variance among different audio samples is also smaller than in previous distillation methods.

\begin{figure*}[t]
    \begin{center}
        \includegraphics[trim=0.0cm 0.0cm 0.0cm 0.0cm,width=0.98\textwidth]{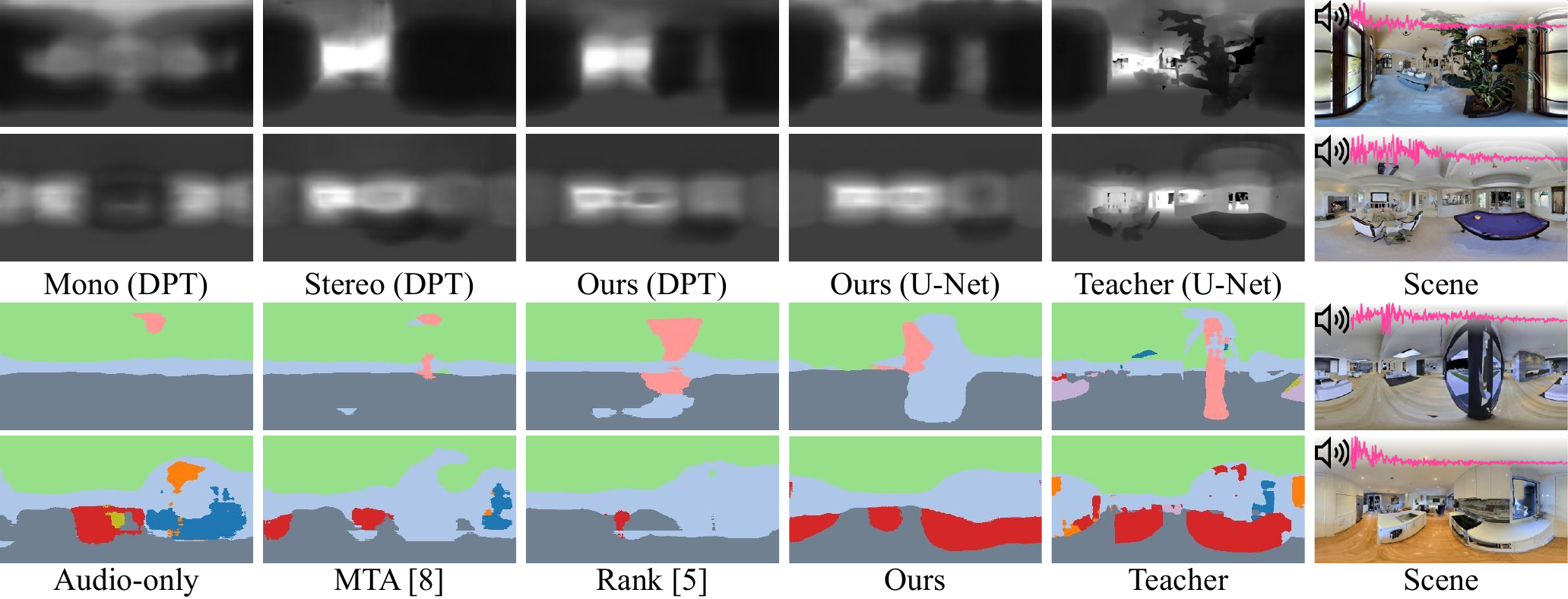}
    \end{center}
    \vspace{-10pt}
    \caption{Qualitative examples of audio-based depth estimation (upper) and semantic segmentation (lower).}
    \label{fig:qual_2d}
    \vspace{-10pt}
\end{figure*}

\textbf{Qualitative Results.}
Fig.~\ref{fig:qual_2d} displays the depth estimation results from binaural audio.
Our approach can precisely measure the depth or structure of the room compared to prior arts.
In some cases, it can even capture smaller objects like a billiards table in a scene from the audio.

\begin{table}[t]
\centering
\begin{tabular}{l|c@{\hspace{1.3\tabcolsep}}c@{\hspace{1.3\tabcolsep}}c@{\hspace{1.3\tabcolsep}}c}
\hline
& pAcc\ua & mAcc\ua & mIoU\ua & 3IoU\ua \\ \hline
Teacher                                & 0.737 & 0.708 & 0.409 & 0.705 \\ \hline
BilinearCoAttn~\cite{irie2022co}       & 0.605 & 0.493 & 0.340 & 0.538 \\  
MM-DistillNet~\cite{valverde2021there} & 0.629 & 0.515 & 0.311 & 0.581 \\  \hline
Pseudo-GT ($\mathcal{L}_p$)~\cite{vasudevan2020semantic} & 0.628 & 0.513 & 0.320 & 0.576 \\
+ MTA~\cite{valverde2021there}         & 0.629 & 0.514 & 0.316 & 0.576 \\
+ Rank~\cite{gan2019self}              & 0.642 & 0.520 & 0.359 & 0.587 \\
+ \textbf{SAM$_\text{Full}$}           & \textbf{0.644} & \textbf{0.526} & \textbf{0.363} & \textbf{0.600} \\ \hline
\end{tabular}
\caption{Comparison of semantic segmentation accuracy on DAPS-Semantic test split.}
\label{tab:semseg}
\vspace{-10pt}
\end{table}

\subsection{Results of Semantic Segmentation}
\label{subsec:semseg}

\subsubsection{Experiment Settings}
We train the audio student model to predict pixel-wise categories of the scene.
Except for the pseudo-GT learning objective $\mathcal{L}_p$, we follow the training recipe explained in Sec.~\ref{subsec:depth}.
As an auxiliary task, we predict the pseudo-GT segmentation with the penultimate layer feature for better performance, as proposed by Zhao \etal~\cite{zhao2017pyramid}.
We train the model with the cross-entropy loss, where the primary and auxiliary loss ratio is 1:0.2.

Since it is virtually not tractable to classify 40+ semantic categories merely from the audio, we opt out classes about tiny objects (\eg, towels) and merge similar classes to establish nine classes for semantic segmentation based on audio.
We report the performance of feature-level distillation methods with U-Net as a backbone.

\textbf{Evaluation Metrics.}
We report the pixel-wise accuracy (pAcc), class-wise mean accuracy (mAcc), and class-wise mean IoU (mIoU) for all pixels with valid labels.
Since it is challenging to label small objects in a scene with audio precisely, we introduce the mean IoU of ceiling, wall, and floor (3IoU) that constitutes a coarse layout of the scene.

\subsubsection{Results and Analyses}

Table~\ref{tab:semseg} summarizes the semantic segmentation accuracy on DAPS-Semantic test split.
Although predicting material properties or a semantic structure from auditory input is challenging, the result suggests that the overall output is acceptably plausible, achieving 87\% of the teacher model's performance on the pAcc metric.
Compared to depth estimation, the ranking-based objective fairly contributes to the distillation performance, which could be related to the classification error ensuring tighter bounds for ranking measures~\cite{chen2009ranking}.
Still, SAM achieves better performance in all metrics, especially in predicting layout-relevant categories, \ie, +4\% compared to Pseudo-GT.

\textbf{Qualitative Examples.}
The last two rows of Fig.~\ref{fig:qual_2d} illustrate the semantic segmentation results.
Our approach can better predict the categories of smaller objects and the layout of the indoor surroundings, even under visually ill-posed scenarios like the windows in the third row.

\begin{figure*}[t]
    \begin{center}
        \includegraphics[trim=0.0cm 0.0cm 0.0cm 0.0cm,width=0.98\textwidth]{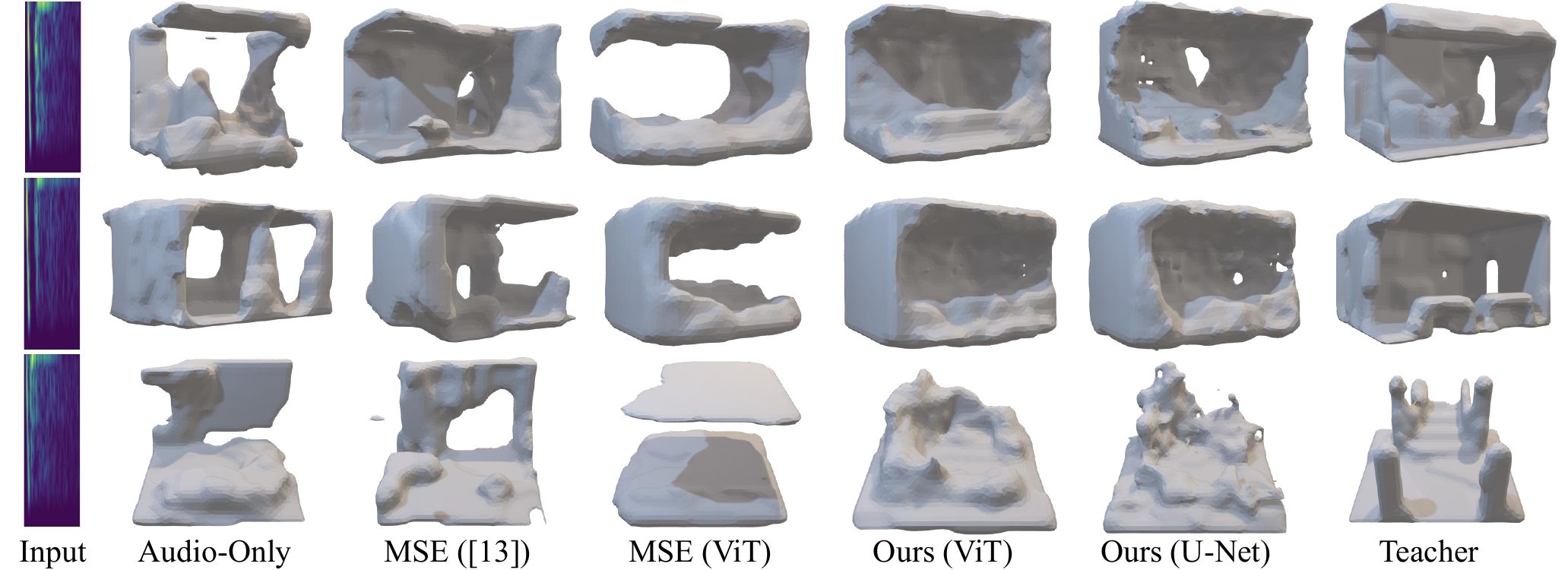}
    \end{center}
    \vspace{-10pt}
    \caption{Qualitative examples of audio-based 3D scene reconstruction.}
    \label{fig:qual_3d}
    \vspace{-10pt}
\end{figure*}
\subsection{Results of 3D Scene Reconstruction}
\label{subsec:3d}

\subsubsection{Experiment Settings}

We reconstruct a 3D scene with audio by means of voxel super-resolution.
Voxel super-resolution aims to reconstruct high-resolution 3D objects using low-resolution voxelized meshes as input~\cite{mescheder2019occupancy}. 
We use a teacher model that maps low ($16^3$) to high-resolution voxel grids ($32^3$) by capturing structural details of 3D shapes for reconstruction.
Despite the difference in dimensions and shapes, the feature maps of the 3D teacher U-Net are utilized to learn the spatial alignment with auditory features, owing to the SAM blocks.

\textbf{Evaluation Metrics.}
Following Peng \etal~\cite{Peng2020ConvolutionalON}, we report IoU, Chamfer-L$_1$ distance, normal consistency (NC), and F1-score. 
We use IoU and F1 to measure the intersection between ground truths and predictions.
Also, we evaluate Chamfer-L$_1$ distance and NC as similarity metrics based on multidimensional point sets and normal displacement vectors, respectively.

\textbf{Baselines.}
Due to a lack of prior research on generating 3D objects from audio, we set up several conceivable baselines for comparison.
First, we interpolate the 2D audio input to 3D to use ConvONet as a backbone.
We report the performance of audio-only models and their variants with feature distillation.
Second, as in the spatial alignment via matching framework, we use the 2D audio input as is and convert intermediate feature maps to match the shape of 3D features.
We use U-Net~\cite{ronneberger2015u} and ViT~\cite{dosovitskiy2021image} as backbones to show that our approach can be applied to various encoder structures, where we include the ranking~\cite{gan2019self} or MTA~\cite{valverde2021there} objectives for cross-modal distillation as baselines.


\begin{table}[t]
\centering
\begin{tabular}{cl|cccc}
\hline
&& IoU\ua & Chamfer\da & NC\ua & F1\ua \\ \hline
\multicolumn{2}{l|}{Teacher~\cite{Peng2020ConvolutionalON}}
& 0.548 & 0.0137     & 0.882 & 0.560 \\ \hline
\multicolumn{2}{l|}{Audio-only$_\text{Mono}$} & 0.126 & 0.0698 & 0.625 & 0.189 \\
\multicolumn{2}{l|}{Audio-only$_\text{Stereo}$}    & 0.136 & 0.0643 & 0.639 & 0.196 \\ \hline
\multirow{3}{*}{\rotatebox[origin=c]{270}{\makecell{\cite{Peng2020ConvolutionalON}}}}
& MSE           & 0.137 & 0.0630 & 0.639 & \textbf{0.203} \\
& Rank~\cite{gan2019self}         & 0.138 & 0.0636 & 0.640 & 0.200 \\
& MTA~\cite{valverde2021there}   & 0.149 & 0.0656 & 0.631 & 0.174 \\ \hline			
\multirow{4}{*}{\rotatebox[origin=c]{270}{\makecell{U-Net \cite{ronneberger2015u}}}}
& MSE                         & 0.150 & 0.0676 & 0.626 & 0.177 \\ 
& Rank~\cite{gan2019self}         & 0.153 & 0.0663 & 0.631 & 0.174 \\
& MTA~\cite{valverde2021there}    & 0.159 & 0.0660 & 0.645 & 0.170 \\ 
& \textbf{SAM$_\text{Full}$}      & \textbf{0.178} & \textbf{0.0555} & \textbf{0.679} & \textbf{0.203} \\ \hline 
\multirow{4}{*}{\rotatebox[origin=c]{270}{\makecell{ViT \cite{ranftl2021vision}}}}
& MSE                           & 0.154 & 0.0626 & 0.656 & 0.183 \\ 
& Rank~\cite{gan2019self}            & 0.147 & 0.0698 & 0.671 & 0.177 \\ 
& MTA~\cite{valverde2021there}       & 0.154 & 0.0650 & 0.646 & 0.187 \\ 
& \textbf{SAM$_\text{Full}$}          & \textbf{0.178} & \textbf{0.0587} & \textbf{0.682} & \textbf{0.204} \\\hline 
\end{tabular}
\captionof{table}{Comparison of 3D scene reconstruction accuracy on DAPS-3D test split.}
\label{tab:occ_full}
\vspace{-10pt}
\end{table}

\subsubsection{Results and Analyses}
Table~\ref{tab:occ_full} reports the 3D scene reconstruction performance on DAPS-3D test split.
Due to task difficulty, the performance gap between the teacher and the student is wider than 2D dense prediction tasks.
Still, our approach improves the IoU score by 40\% compared to audio-only models.
Instead of forcefully converting the audio input representation, reducing the feature distance while keeping the audio input intact generally performs better.
Lower Chamfer-L$_1$ scores of our approach, \ie, an 18\% reduction for the U-Net backbone, suggest that SAM facilitates the generation of points that are significantly closer to the ground truth.

\textbf{Qualitative Examples.}
Fig.~\ref{fig:qual_3d} visualizes our audio-based 3D scene reconstruction results.
In the absence of visual cues, our approach accurately predicts the closed walls in a scene, even capturing details like holes (\eg, doors or windows) and furniture.
The substantial gap of quality between ours and prior arts in an open space (the last row of Fig.~\ref{fig:qual_3d}) stresses the importance of our distillation framework for dense prediction of 3D surroundings.

\section{Conclusion}
\label{sec:conclusion}

We addressed the audio-based dense prediction of indoor surroundings in 2D and 3D for the first time, addressing the challenges in vision-to-audio knowledge distillation: the discrepancy between the two modalities.
To this end, we presented a novel spatial alignment via matching (SAM) distillation framework, accounting for local correspondence of multi-scale features with input shape inconsistency.
In experiments in a newly collected DAPS dataset, our distillation framework consistently improves the performance across multiple tasks ranging from 2D to 3D with various architectures as backbones.
Qualitative results indicate that our approach better captures fine-grained information about the scene from the auditory input compared to prior arts.


\textbf{Acknowledgement}. 
This work was supported by LG AI Research, National Research Foundation of Korea (NRF) grant (No.2023R1A2C2005573) and Institute of Information \& Communications Technology Planning \& Evaluation (IITP) grant (No.2022-0-00156, 2019-0-01082, 2021-0-01343) funded by the Korea government (MSIT). Gunhee Kim is the corresponding author.

{\small
\bibliographystyle{unsrt} 
\bibliography{iccv23-aesop}
}

\end{document}